\documentclass[letterpaper, 10 pt, conference]{ieeeconf}  
\usepackage{graphicx}
\usepackage{stfloats}
\usepackage{float}
\usepackage{caption}
\usepackage{cuted}
\usepackage{amsmath}
\usepackage{mathtools}
\usepackage{amssymb}
\usepackage{booktabs} 
\usepackage{multirow}
\usepackage{makecell}

\usepackage{tabularx}
\usepackage[table]{xcolor}
\usepackage{array}
\definecolor{cStack}{RGB}{232,242,255}
\definecolor{cPress}{RGB}{255,241,230}
\definecolor{cHammer}{RGB}{236,255,236}
\definecolor{cSafe}{RGB}{250,240,255}
\definecolor{cOpen}{RGB}{255,252,230}
\definecolor{cPour}{RGB}{235,245,245}
\definecolor{cOverall}{RGB}{240,240,240}
\newcommand{\method}{\texttt{EmboAlign}}

\IEEEoverridecommandlockouts                              

\usepackage[T1]{fontenc}

\title{\LARGE \bf
\method{}: Aligning Video Generation with Compositional Constraints for Zero-Shot Manipulation
}
\newcommand{\rev}[1]{#1}

\author{
  Gehao Zhang$^{1,*}$, Zhenyang Ni$^{1,*}$, Payal Mohapatra$^{1}$, Han Liu$^{1}$, Ruohan Zhang$^{1,2}$, Qi Zhu$^{1}$%
  \thanks{$^{*}$ Equal contribution.}\\
  {\normalsize $^{1}$Northwestern University \quad $^{2}$Stanford University}
}

\begin{document}

\maketitle

\thispagestyle{empty}
\pagestyle{empty}


\begin{abstract}
Video generative models (VGMs) pretrained on large-scale internet data can produce temporally coherent rollout videos that capture rich object dynamics, offering a compelling foundation for zero-shot robotic manipulation. However, VGMs often produce physically implausible rollouts, and converting their pixel-space motion into robot actions through geometric retargeting further introduces cumulative errors from imperfect depth estimation and keypoint tracking.
To address these challenges, we present \method{}, a data-free framework that aligns VGM outputs with compositional constraints generated by vision-language models (VLMs) at inference time. The key insight is that VLMs offer a capability complementary to VGMs: structured spatial reasoning that can identify the physical constraints critical to the success and safety of manipulation execution. Given a language instruction, \method{} uses a VLM to automatically extract a set of compositional constraints capturing task-specific requirements, which are then applied at two stages: (1) constraint-guided rollout selection, which scores and filters a batch of VGM rollouts to retain the most physically plausible candidate, and (2) constraint-based trajectory optimization, which uses the selected rollout as initialization and refines the robot trajectory under the same constraint set to correct retargeting errors. We evaluate \method{} on six real-robot manipulation tasks requiring precise, constraint-sensitive execution, improving the overall success rate by \rev{43.3 percentage points} over the strongest baseline without any task-specific training data.
\end{abstract}


\section{INTRODUCTION}
Generalizable robotic manipulation remains a central challenge in robotics, as real-world tasks demand policies that transfer across diverse objects, scenes, and instructions without costly task-specific retraining~\cite{zitkovich2023rt,kim2024openvla,black2026pi0visionlanguageactionflowmodel,shridhar2021cliportpathwaysroboticmanipulation,brohan2023rt1roboticstransformerrealworld}.
Recent advances in video generative models (VGMs) trained on large-scale internet data have opened a promising path toward this goal: conditioned on an initial observation and a language instruction, modern VGMs can produce temporally coherent rollout videos that capture rich object dynamics, contact evolution, and goal-directed behavior~\cite{du2023learning,chen2024lvp,rigvid,ni2026creflowcorrectivereflowsparsereward}.
This has motivated a growing line of work on \emph{video-based manipulation}~\cite{du2023learning,chen2024lvp,zhang2024videoagent,wen2023atm,rigvid,dream2flow,mao2024physworld}, which generates a video plan from the task instruction and current observation, then retargets the predicted motion into robot actions via depth estimation or pose tracking.


Despite this progress, video-based manipulation pipelines suffer from two compounding failure modes that are particularly acute for precise manipulation.
First, VGMs frequently exhibit physical hallucinations (object interpenetration, non-conservative motion, or prompt-following drift) because they are trained on large-scale, diverse video corpora where physically-grounded interaction data remains scarce. Second, converting pixel-space video motion into robot actions through geometric retargeting introduces cumulative errors from imperfect depth estimation and keypoint tracking, causing large execution failures even from visually plausible rollouts.
A key observation is that successful manipulation inherently requires satisfying compositional constraints, including spatial relations (e.g., ``block A must be placed on block B''), kinematic requirements (e.g., ``approach the object from above''), and safety conditions (e.g., ``avoid the obstacle''). Yet current VGM-based pipelines lack mechanisms to enforce them, leading to task failure or even safety hazards.

\begin{figure}[t]
\centering
\includegraphics[width=0.85\linewidth]{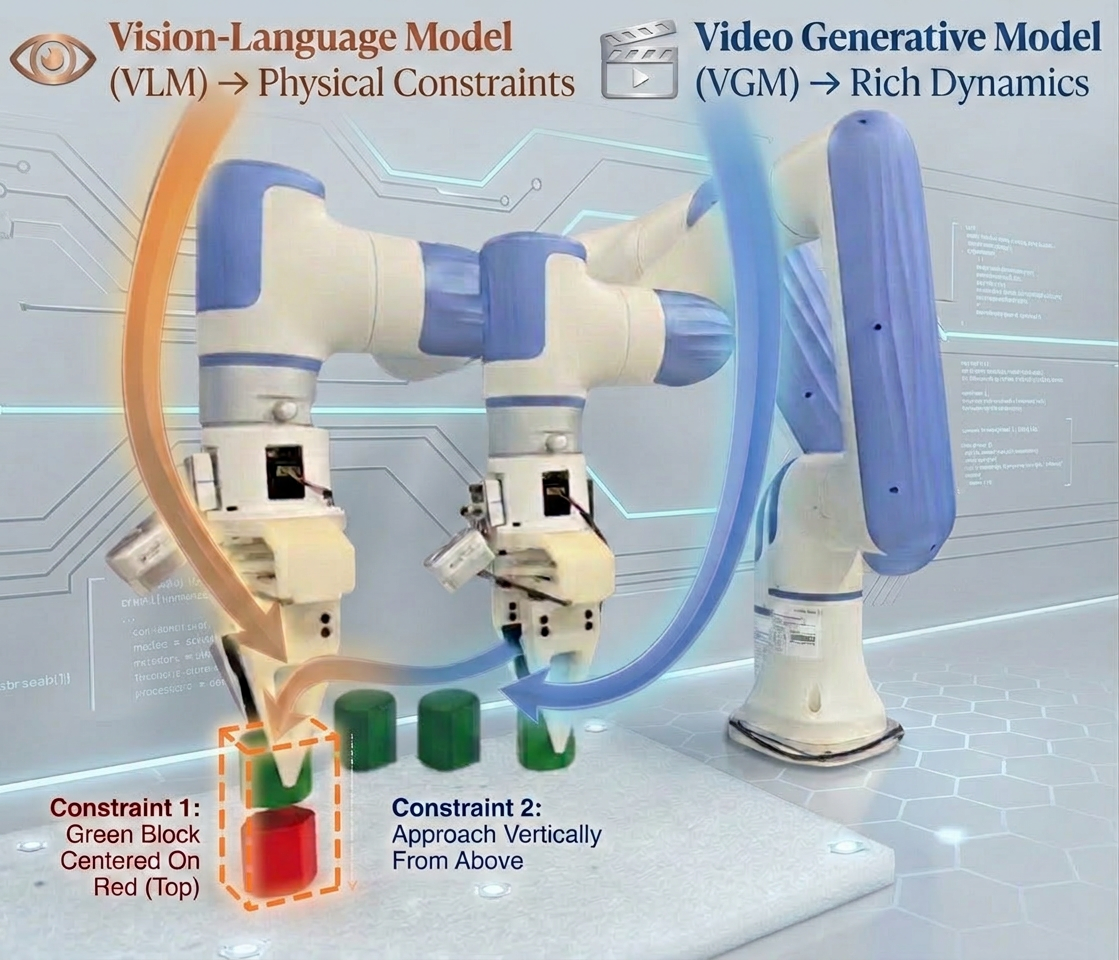}
\caption{Video Generation Models can zero-shot generate rich motion priors for manipulation tasks, but hallucinations and retargeting errors may prevent these from translating into correct robot actions. We propose to use VLM-derived compositional constraints (e.g., $c_1$: placement alignment, $c_2$: top-down approach) to align VGM outputs at both the video selection and trajectory optimization stages, bridging the gap between generative motion diversity and the physical precision that real-world manipulation demands.}
\label{fig:teaser}
\vspace{-5mm}
\end{figure}

To address both failure modes within a unified framework, we propose \method{}, a \emph{compositional constraint alignment} pipeline that uses a vision-language model (VLM) to automatically extract task-specific constraints from language instructions and employs them at two critical stages of the video-to-action pipeline, as illustrated in Fig.~\ref{fig:teaser}.
The core insight is that the capabilities of VLMs and VGMs are \emph{complementary}: VGMs provide generative diversity and rich motion priors from large-scale pretraining, while VLMs bring structured physical reasoning and semantic grounding that VGMs lack.
Specifically, given a manipulation instruction, we first use a VLM to decompose the task into a set of compositional physical and relational constraints (e.g., ``the gripper must approach from above,'' ``the object must not exceed a velocity threshold'').
These constraints serve two roles: \textbf{(1) Constraint-guided rollout selection}: we sample a large batch of rollout videos from the VGM and use the VLM-derived constraints as a scoring function to filter and select rollouts that are most physically plausible and semantically consistent with the instruction; \textbf{(2) Constraint-based trajectory optimization}: the selected rollout is used to initialize a constrained trajectory optimization procedure that retargets the video motion into feasible robot joint actions, using the same constraint set as hard or soft optimization objectives to prevent local minima and correct retargeting errors in real time.
This two-stage use of constraints corrects hallucinations of the VGM at the planning level and improves the precision of action retargeting at the execution level.

We evaluate \method{} on six real-robot manipulation tasks, each requiring precise, constraint-sensitive execution (e.g., block stacking, tool use, safety-constrained placement).
Our method improves the overall success rate by \rev{43.3 percentage points} over
the strongest baseline without any task-specific training data, demonstrating
that compositional constraint alignment is a principled and effective approach
to bridging the gap between internet-pretrained VGMs and the physical demands
of real-world manipulation.

In summary, the main contributions of this work include:
\begin{itemize}
    \item We introduce \method{}, a novel framework that aligns video generative model rollouts with manipulation task requirements through compositional constraints, enabling precise and safe zero-shot execution. 
	\item We design a two-stage constraint alignment mechanism: constraint-guided rollout selection to filter physically implausible VGM samples, followed by constraint-based trajectory optimization to correct retargeting errors, addressing the inherent limitations of VGM-based manipulation pipelines within a unified framework.
    \item We validate \method{} on six real-robot manipulation tasks, improving the
    overall success rate by \rev{43.3 percentage points} over the strongest baseline
without any task-specific training data.
\end{itemize}

\section{RELATED WORK}

\subsection{Video Generative Models for Manipulation}
Video generative models (VGMs) have emerged as a promising foundation for robotic manipulation~\cite{yang2024videolanguage,mccarthy2025survey}.
A prominent paradigm casts decision-making as conditional video generation: UniPi~\cite{du2023learning} pioneered text-conditioned video synthesis with inverse dynamics, followed by works that scale with larger foundation models~\cite{chen2024lvp,zhang2024videoagent}, compose hierarchical language--video--action planners~\cite{ajay2024hip}, generate human demonstrations as video guidance~\cite{gen2act,liang2024dreamitate}, or synthesize visual subgoals and training trajectories from video world models~\cite{liang2024envision,jang2025dreamgen}.
Rather than modifying the VGM, a complementary line extracts actionable motion signals from its outputs via dense correspondences~\cite{xu2024flowcrossdomainmanipulationinterface,chen2025g3flowgenerative3dsemantic}, 6D pose tracking~\cite{rigvid}, 3D flow through depth estimation and point tracking~\cite{li2024novaflow,dream2flow}, or object-centric reinforcement learning in simulation~\cite{mao2024physworld}.
Large-scale video data has also been used to pre-train representations that transfer to downstream control, including GPT-style joint action--video predictors~\cite{wu2023gr1}, latent world models trained on action-free videos~\cite{seo2022apv,wu2023contextwm}, spatiotemporal predictive encoders~\cite{yang2024stp}, video diffusion feature extractors~\cite{hu2024vpp}, and unified video--action models~\cite{li2025uvam}.
Beyond planning and representation, video models have been repurposed as reward functions for reinforcement learning via prediction likelihoods~\cite{escontrela2023viper}, conditional diffusion entropy~\cite{huang2024diffusionreward}, or video discrimination~\cite{chen2021dvd}, and as interactive world simulators that model environment dynamics from video~\cite{yang2023unisim,bruce2024genie,zhou2024robodreamer}.
A central challenge across these approaches is that video-generated plans often violate physical constraints, causing failures when mapped to actions.
GVP-WM~\cite{ziakas2025gvpwm} addresses this by projecting video plans onto feasible latent trajectories via optimization in a learned world model, and Luo and 
Du~\cite{luo2024groot} ground video models through goal-conditioned exploration.
Our work tackles this challenge from a complementary angle: \method{} uses VLM-derived compositional constraints to select physically plausible rollouts and refine retargeted trajectories, requiring no pre-trained world model or test-time dynamics optimization.

\subsection{Constraint-Based Manipulation}
Compositional constraint formulations provide an expressive interface for specifying the spatial, temporal, and interaction requirements of manipulation tasks.  
ReKep~\cite{rekep} uses a VLM to automatically generate relational keypoint constraints as Python cost functions and solves for robot actions via hierarchical constrained optimization, enabling diverse multi-stage and bimanual behaviors from language instructions alone.
Code-as-Monitor~\cite{yang2024cam} extends this paradigm to failure detection, using VLM-generated code to evaluate spatio-temporal constraint satisfaction both reactively and proactively.
SafeBimanual~\cite{safebimanual} applies constraint-based thinking to safety in dual-arm manipulation, introducing a test-time trajectory optimization framework that adds safety constraints to a diffusion-based policy.
\rev{Beyond manipulation, safe reinforcement learning enforces safety constraints on learning-enabled agents~\cite{wang2023enforcing,zhan2024statewise}, and safety-related constraints have been evaluated for foundation model-based embodied agents~\cite{wang2023empowering,zhan2025sentinel}.}
Maestro~\cite{han2024maestro} uses a VLM coding agent to compose specialized perception, planning, and control modules into a programmatic policy~\cite{huang2023voxposercomposable3dvalue,liang2023codepolicieslanguagemodel}.
VLMPC~\cite{vlmpc} integrates VLM-derived constraints into a model predictive control framework, using language-grounded cost functions to guide action sampling and selection.
However, these constraint-based methods typically rely on local optimizers that are sensitive to initialization; without a good initial trajectory, the solver can converge to poor local optima or fail entirely, limiting their applicability to complex, long-horizon tasks.
\method{} addresses this by using VGM-generated rollouts as trajectory initializations, combining the motion diversity of video models with the physical precision of constraint-based optimization.

\section{METHOD}
\label{sec:method}

\begin{figure*}[t]
  \centering
  \includegraphics[width=\linewidth]{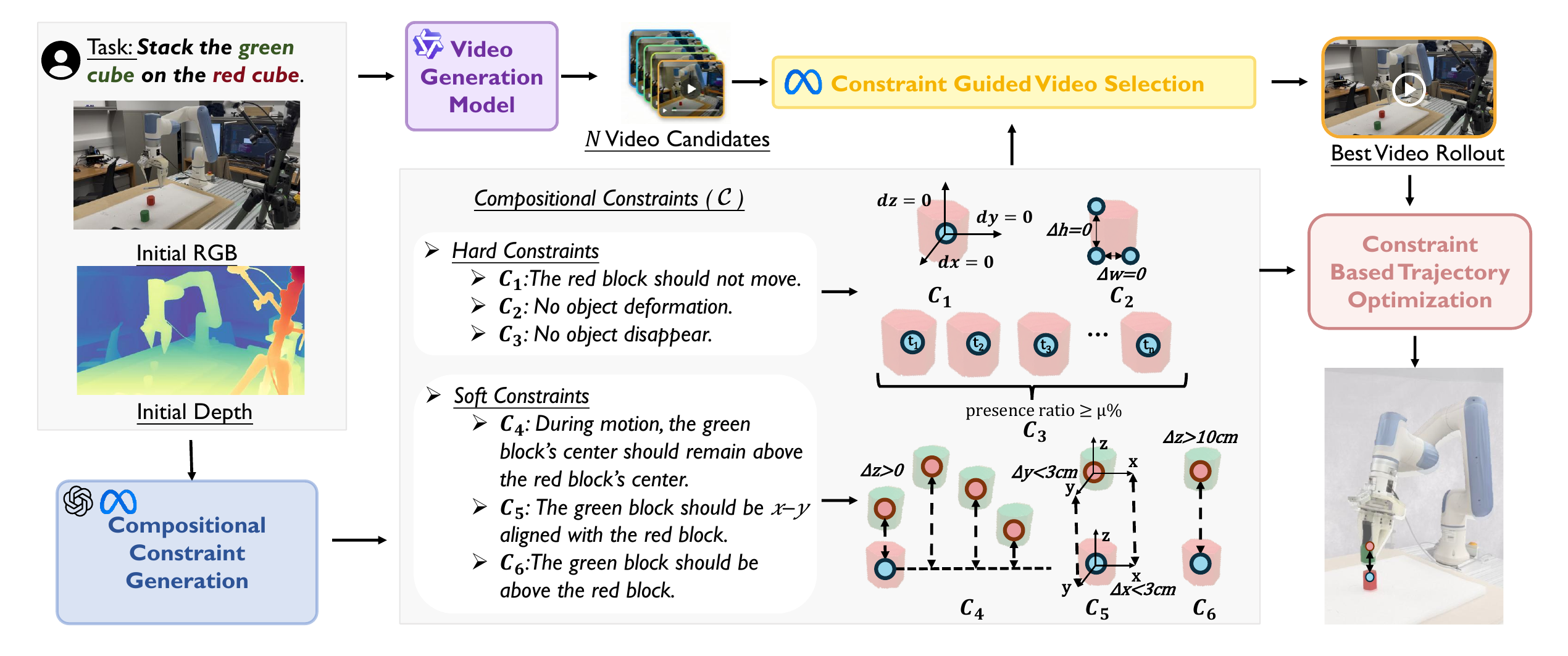}
  \caption{\textbf{\method{} pipeline.} Given a language instruction and RGB--D observations, a VLM generates compositional constraints while a VGM produces candidate rollout videos. A latent world model ranks rollouts by physical plausibility, then the constraint set filters candidates in descending-score order. The top valid rollout is retargeted into an end-effector trajectory and optimized under the same constraints for real-world execution.}
  \label{fig:pipeline}
  \vspace{-5mm}
\end{figure*}

\subsection{Problem Formulation}
\label{sec:method:setup}

We consider zero-shot robotic manipulation given an initial RGB--D observation $\mathbf{o} = (\mathbf{I}, \mathbf{D})$ and a language instruction $\ell$.
The objective is to produce a feasible end-effector trajectory $\boldsymbol{\xi}_{1:T} = (\boldsymbol{\xi}_1, \ldots, \boldsymbol{\xi}_T)$, where each $\boldsymbol{\xi}_t \in \mathrm{SE}(3)$ (the group of rigid-body transformations comprising 3D rotation and translation) denotes a 6-DoF end-effector pose, such that the robot completes the instructed task while satisfying all relevant physical and geometric constraints.
We represent each task-relevant object by a sparse set of 3D keypoints $\mathbf{k} \in \mathbb{R}^{K \times 3}$, where $K$ is the total number of keypoints across all objects in the scene.
Keypoints provide a general, object-agnostic geometric representation that can capture spatial relations, contact conditions, and placement requirements across diverse tasks; detailed extraction is described in Sec.~\ref{sec:method:constraint}.

Our framework proceeds as follows, as illustrated in Fig.~\ref{fig:pipeline}.
First, a vision--language model (VLM) parses the observation and instruction into a set of compositional constraints:
\begin{equation}
  \mathcal{C} = f_{\text{vlm}}(\mathbf{o},\, \ell),
  \tag{1a}
  \label{eq:constraint_gen}
\end{equation}
where each constraint $c \in \mathcal{C}$ is a scalar-valued function over the keypoint configuration, i.e., $c\colon \mathbb{R}^{K \times 3} \to \mathbb{R}$, with $c(\mathbf{k}) \leq 0$ indicating satisfaction (Sec.~\ref{sec:method:constraint}).
Second, $N$ candidate rollout videos are sampled from a pretrained video generative model (VGM) and the most plausible, constraint-consistent candidate is selected:
\begin{equation}
\begin{aligned}
  &V_i \sim p_{\text{vgm}}(\cdot \mid \mathbf{o},\, \ell),\; i = 1, \ldots, N, \\
  &V^{*} = \operatorname{Select}\!\bigl(\{V_i\}_{i=1}^{N};\, \mathcal{C}\bigr),
\end{aligned}
  \tag{1b}
  \label{eq:video_select}
\end{equation}

where the selection procedure jointly evaluates visual coherence and spatial constraint satisfaction (Sec.~\ref{sec:method:selection}).
Third, the selected rollout $V^{*}$ is converted into an initial end-effector trajectory via grasp-conditioned retargeting:
\begin{equation}
  \boldsymbol{\xi}^{(0)}_{1:T} = g_{\text{retarget}}\!\bigl(V^{*},\, \mathbf{o}\bigr),
  \tag{1c}
  \label{eq:retarget}
\end{equation}
where $g_{\text{retarget}}$ lifts video-space object motion into a robot-executable pose sequence.
Fourth, this trajectory is refined under the same constraint set $\mathcal{C}$ to correct retargeting errors:
\begin{equation}
  \boldsymbol{\xi}^{*}_{1:T}
  = \arg\min_{\boldsymbol{\xi}_{1:T}}\; J\!\bigl(\boldsymbol{\xi}_{1:T};\, \mathcal{C}\bigr),
  \quad \text{initialized from } \boldsymbol{\xi}^{(0)}_{1:T},
  \tag{1d}
  \label{eq:traj_opt}
\end{equation}
where $J$ penalizes constraint violations along the trajectory (Sec.~\ref{sec:method:traj_opt}).
The remainder of this section details each stage.
\setcounter{equation}{1}

\subsection{Compositional Constraint Generation}
\label{sec:method:constraint}

Compositional constraints have emerged as a powerful interface for specifying manipulation tasks, and recent work has explored a variety of object representations for formulating them, including 6D object poses~\cite{omnimanip}, annotated functional axes~\cite{robotwin}, geometric primitives such as points, lines, and surfaces~\cite{yang2024cam}, and sparse keypoints~\cite{rekep,gensim2,safebimanual}.
We adopt the keypoint-based formulation, as keypoints provide a general, object-agnostic representation that naturally captures spatial relations, contact conditions, and placement requirements across diverse manipulation tasks without requiring object-specific models or axis annotations.

\begin{figure*}[t]
\centering
\includegraphics[width=\linewidth]{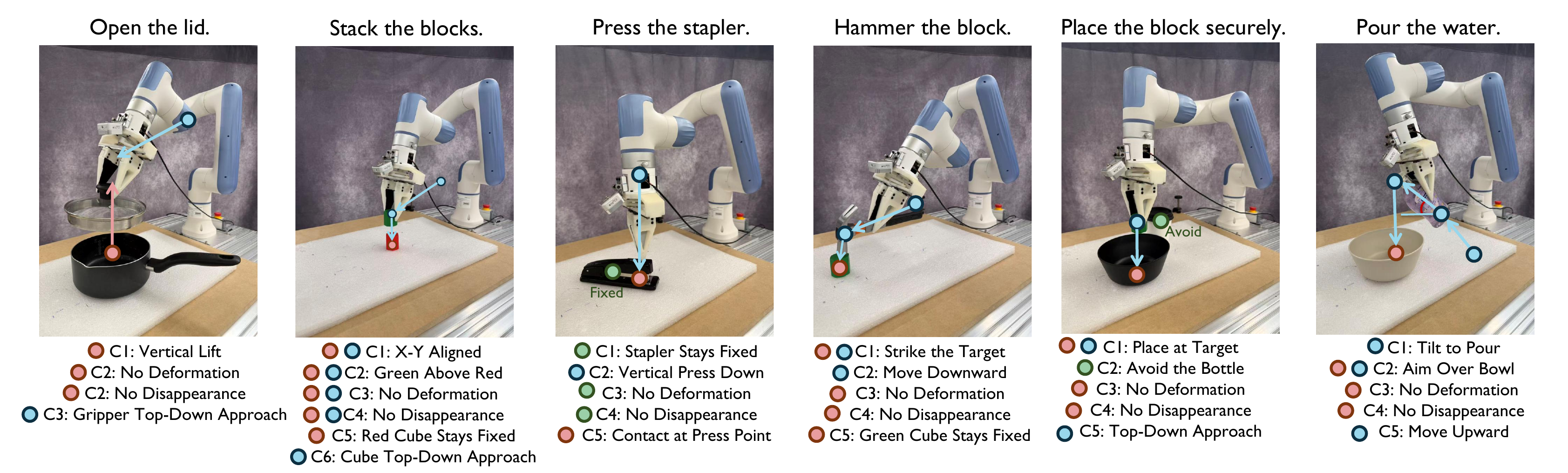}
\caption{Optimization constraints for real-robot evaluation. For each of the six manipulation tasks, a VLM automatically generates a set of constraints encoding spatial, kinematic, and safety requirements. These constraints serve as optimization objectives for trajectory refinement during execution.}
\label{fig:task_constraints}
\vspace{-5mm}
\end{figure*}

Concretely, we apply Segment Anything~\cite{sam,liu2023grounding} to the initial RGB image, $\mathbf{I}$, to obtain instance masks $\{M_e\}_{e \in \mathcal{E}}$ for task-relevant entities $\mathcal{E}$, and sample a sparse set of 2D keypoints $\mathcal{P}_e = \{\mathbf{p}_{e,j}\}_{j=1}^{J_e}$ from each mask (interior samples and geometric extrema).
We then render $\mathbf{I}$ with the indexed keypoints overlaid and prompt a VLM to produce the constraint set $\mathcal{C} = f_{\text{vlm}}(\mathbf{o}, \ell)$.
Each constraint $c \in \mathcal{C}$ is a Python function that maps a 3D keypoint configuration $\mathbf{k} \in \mathbb{R}^{K \times 3}$ to a scalar cost, where $c(\mathbf{k}) \leq 0$ indicates satisfaction~\cite{rekep,safebimanual}.
The constraint set may include both goal-state conditions (e.g., ``the block is centered on the target'') and process-level requirements (e.g., ``the gripper approaches from above''); our framework treats all constraints uniformly in both downstream stages. Fig.~\ref{fig:task_constraints} shows the generated constraints for all six evaluation tasks.

Given a 3D keypoint trajectory $\mathcal{K} = \{\mathbf{k}_t\}_{t=1}^{T}$, the aggregate constraint violation is:
\begin{equation}
  \mathrm{cost}_{\mathcal{C}}(\mathcal{K})
  = \sum_{c \in \mathcal{C}} \sum_{t=1}^{T} \bigl[\max\!\bigl(0,\; c(\mathbf{k}_t)\bigr)\bigr]^2.
  \label{eq:cost}
\end{equation}
This cost is used in both the video selection stage (Sec.~\ref{sec:method:selection}) and the trajectory optimization stage (Sec.~\ref{sec:method:traj_opt}).

\subsection{Constraint-Guided Video Selection}
\label{sec:method:selection}

Given $(\mathbf{o}, \ell)$, we sample a batch of $N$ candidate rollouts $\mathcal{V} = \{V_i\}_{i=1}^{N}$ from a pretrained VGM.
To select a rollout that is both visually coherent and physically consistent with the task, we evaluate each candidate against two complementary criteria: a \emph{visual plausibility} score that captures low-level temporal coherence, and a \emph{spatial constraint} score that verifies whether the depicted motion satisfies the task requirements encoded in $\mathcal{C}$.

\paragraph{Visual plausibility}
We employ V-JEPA-2~\cite{vjepa}, a self-supervised latent world model, as a learned metric for visual plausibility.
The key intuition is that latent world models learn to predict future video representations in a compressed feature space, focusing on underlying physical dynamics rather than pixel-level details.
When a generated rollout is physically plausible, V-JEPA-2's predictions closely match the observed future; when the rollout contains hallucinations (e.g., object morphing, non-physical motion), the prediction diverges.
We leverage this divergence as a reward signal.
Concretely, for a rollout $V = \{v_t\}_{t=1}^{T}$, we slide a context prediction window across the frame sequence.
At each anchor position $s$, the V-JEPA-2 encoder $E_\theta$ processes $C$ context frames and the predictor $P_\phi$ forecasts latent representations for the subsequent $M$ frames.
The visual plausibility score is defined as the mean cosine discrepancy between predicted and observed latent representations:
\begin{equation}
  s_{\text{vis}}(V) = \frac{1}{|\mathcal{S}|} \sum_{s \in \mathcal{S}}
  \left(
    1 - \frac{\tilde{\mathbf{z}}_s^\top \mathbf{z}_s}
    {\|\tilde{\mathbf{z}}_s\|_2 \,\|\mathbf{z}_s\|_2}
  \right),
  \label{eq:vjepa_score}
\end{equation}
where $\tilde{\mathbf{z}}_s = P_\phi\!\bigl(\Delta_m,\, E_\theta(v_{s-C+1:s})\bigr)$ is the predicted representation and $\mathbf{z}_s = E_\theta(v_{s-C+1:s+M})$ is the encoded observation.
Rollouts with lower $s_{\text{vis}}$ exhibit more predictable dynamics and fewer visual artifacts, as the latent world model's predictions align well with physically coherent futures.

\paragraph{Spatial constraint satisfaction}
To evaluate whether a candidate rollout satisfies the task-level constraints in $\mathcal{C}$, we transform its 2D keypoints into 3D trajectories.
Concretely, we:
(i) track the initial keypoints $\{\mathcal{P}_e\}$ through the rollout using CoTracker~\cite{cotracker}, obtaining 2D trajectories $\{\boldsymbol{\pi}_{e,j,t} \in \mathbb{R}^2\}$;
(ii) estimate per-frame depth maps $\{\hat{D}_t\}$ with a monocular video depth estimator (RollingDepth~\cite{rollingdepth});
(iii) resolve the global scale--shift ambiguity by fitting an affine transform $(\alpha, \beta)$ such that $\alpha \hat{D}_1 + \beta \approx \mathbf{D}$ on valid pixels, yielding calibrated depth maps $D_t = \alpha \hat{D}_t + \beta$; and
(iv) back-project the tracked 2D keypoints into 3D using the calibrated depth and known camera intrinsics, producing object-centric 3D keypoint trajectories $\mathcal{K}_i = \{\mathbf{k}_{e,j,t} \in \mathbb{R}^3\}$.
The spatial constraint score for rollout $V_i$ is then computed as:
\begin{equation}
  s_{\text{spatial}}(V_i;\, \mathcal{C}) = \mathrm{cost}_{\mathcal{C}}(\mathcal{K}_i),
  \label{eq:spatial_score}
\end{equation}
following Eq.~\eqref{eq:cost}.

\paragraph{Joint selection}
Since the 3D reconstruction required for spatial evaluation is computationally expensive, we use a sequential evaluation scheme to avoid reconstructing all $N$ candidates.
We first rank all candidates by their visual plausibility score $s_{\text{vis}}$ in ascending order (most plausible first) and then evaluate $s_{\text{spatial}}$ one by one in this order, accepting the first rollout whose spatial constraint cost falls below a threshold $\epsilon$:
\begin{equation}
  V^{*} = V_{(j^{*})}, \quad
  j^{*} = \min\bigl\{ j :\; s_{\text{spatial}}\bigl(V_{(j)};\, \mathcal{C}\bigr) \leq \epsilon \bigr\}.
  \label{eq:joint_select}
\end{equation}
This strategy prioritizes visually coherent candidates while guaranteeing that the selected rollout satisfies the task-level spatial constraints. 


\subsection{Constraint-Based Trajectory Optimization}
\label{sec:method:traj_opt}

The selected rollout $V^{*}$ encodes task-relevant object motion in pixel space.
We convert its 3D keypoint trajectories $\mathcal{K}^{*}$ into an end-effector trajectory through grasp-conditioned retargeting, and then refine it under the same constraint set $\mathcal{C}$ used during video selection.

\paragraph{Grasp estimation}
We apply AnyGrasp~\cite{anygrasp} to the initial scene point cloud to predict a set of stable grasp candidates on the target object.
To improve robustness under partial occlusion, we additionally reconstruct a 3D object model using SAM~3D~\cite{sam3d}, fuse it into the scene point cloud, and re-sample and validate grasp candidates on this augmented representation.
The highest-scoring grasp defines the gripper--object transform $\mathbf{T}_{\text{grasp}} \in \mathrm{SE}(3)$.

\paragraph{Motion retargeting}
Following~\cite{rigvid,dream2flow}, we assume a fixed gripper--object transform $\mathbf{T}_{\text{grasp}}$ throughout the manipulation phase.
Under this assumption, the object-centric keypoint motion in $\mathcal{K}^{*}$ induces a corresponding end-effector motion: at each timestep $t$, we recover the object pose $\mathbf{T}^{\text{obj}}_t \in \mathrm{SE}(3)$ by fitting a rigid transform to the keypoint correspondences between frame $t$ and the initial frame, and compute the end-effector pose as $\boldsymbol{\xi}^{(0)}_t = \mathbf{T}^{\text{obj}}_t \cdot \mathbf{T}_{\text{grasp}}$.

\paragraph{Trajectory optimization}
The retargeted trajectory $\boldsymbol{\xi}^{(0)}_{1:T}$ inevitably accumulates errors from imperfect depth estimation, keypoint tracking noise, and rigid-body fitting.
We correct these by solving the following nonlinear program:
\begin{equation}
  \boldsymbol{\xi}^{*}_{1:T}
  = \arg\min_{\boldsymbol{\xi}_{1:T}}\;
  \sum_{c \in \mathcal{C}} \sum_{t=1}^{T}
  \bigl[\max\!\bigl(0,\, c(\mathbf{k}_t)\bigr)\bigr]^2
  + \lambda \sum_{t=1}^{T} \bigl\|\boldsymbol{\xi}_t - \boldsymbol{\xi}^{(0)}_t\bigr\|^2,
  \label{eq:traj_opt_detail}
\end{equation}
where $\mathbf{k}_t$ denotes the 3D keypoint configuration induced by the end-effector pose $\boldsymbol{\xi}_t$ through the fixed gripper--object transform.
The objective seeks a trajectory that stays as close as possible to the VGM-generated motion prior while satisfying all physical constraints; the first term penalizes constraint violations and the second term preserves fidelity to the video rollout.
The coefficient $\lambda$ controls the trade-off between the two objectives.
We solve this program using Sequential Least Squares Programming (SLSQP)~\cite{slsqp} initialized from $\boldsymbol{\xi}^{(0)}_{1:T}$, with decision variables normalized to $[0, 1]$.
The optimized trajectory $\boldsymbol{\xi}^{*}_{1:T}$ is executed by the robot controller as a 6-DoF end-effector pose sequence.

\section{EXPERIMENTS} 

We evaluate \method{} on six real-robot manipulation tasks that require precise, constraint-sensitive execution.
We compare against a constraint-only baseline and a video-only baseline, ablate key components, and analyze failure modes.

\begin{figure}[b]
  \vspace{-7mm}
  \centering
  \includegraphics[width=0.85\linewidth]{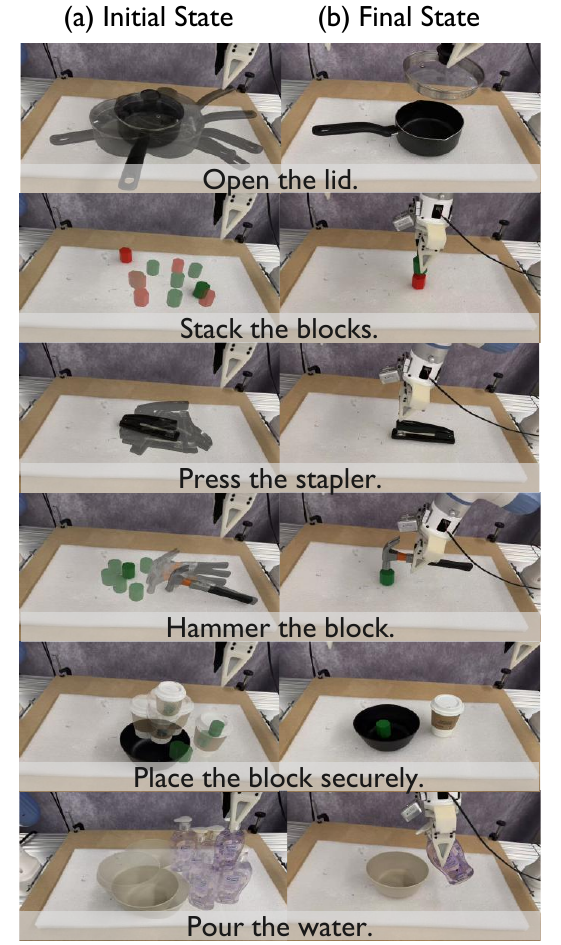}
  \caption{\textbf{Task examples.} Example scenes for the real-robot evaluation tasks.}
  \label{fig:tasks}
\end{figure}
\subsection{Tasks}
We evaluate \method{} in real-world experiments on a Dobot Nova2 robot. As shown in Fig.~\ref{fig:tasks}, our task suite spans diverse objects and interaction modes:
\begin{enumerate}
    \item \textbf{Open the Lid:} A container with a lid is placed on the workspace. A trial is a success if the robot separates the lid from the container (or reaches a target opening displacement) by the end of execution.
    \item \textbf{Stack the Blocks:} A green block and a red block are placed on the workspace. A trial is a success if the green block is placed stably on top of the red block at the end of execution.
    \item \textbf{Press the Stapler:} A stapler is placed on the workspace. A trial is a success if the robot presses the stapler to a target pressed state by the end of execution.
    \item \textbf{Hammer the Block:} A hammer and a target block are placed on the workspace. A trial is a success if the robot uses the hammer to strike the block and achieves a target displacement.
    \item \textbf{Place the Block Securely:} The robot must place an object at a target location while avoiding a water bottle positioned in the workspace. A trial is a success if the object reaches the target without contacting or tipping the bottle.
    \item \textbf{Pour the Water:} A container and a receiving bowl are placed on
the workspace. A trial is a success if the robot achieves a visibly successful
pour into the target bowl by the end of execution.
\end{enumerate}


\subsection{Quantitative Results}

%

\begin{table}[t]
\centering
\caption{\textbf{Task Success Rate on Real Robot.} Success counts out of 10 trials per task.}
\label{tab:success_rekep_novaflow_ours}
\renewcommand{\arraystretch}{1.15}
\setlength{\tabcolsep}{3pt}
\small
\begin{tabularx}{\linewidth}{@{}
    >{\raggedright\arraybackslash}m{1.4cm}
    >{\centering\arraybackslash}X
    >{\centering\arraybackslash}X
    >{\centering\arraybackslash}X
    >{\centering\arraybackslash}X
    >{\centering\arraybackslash}X
    >{\centering\arraybackslash}X
    >{\centering\arraybackslash}m{0.7cm}
@{}}
\toprule
\textbf{Method} &
\textbf{Stack} &
\textbf{Press} &
\textbf{Ham.} &
\textbf{Place} &
\textbf{Open} &
\textbf{Pour} &
\textbf{Avg.} \\
\midrule
ReKep     & 3/10 & 2/10 & 1/10 & 1/10 & 4/10 & 2/10 & 21.7\% \\
NovaFlow  & 2/10 & 0/10 & 1/10 & 4/10 & 4/10 & 4/10 & 25.0\% \\
\midrule
\textbf{Ours}       & \textbf{7/10} & \textbf{8/10} & \textbf{4/10} & \textbf{8/10} & \textbf{7/10} & \textbf{7/10} & \textbf{68.3\%} \\
\bottomrule
\end{tabularx}
  \vspace{-5mm}
\end{table}

We compare \method{} against two baselines: ReKep~\cite{rekep}, a constraint-only method that plans directly from VLM-generated keypoint constraints without video guidance, and NovaFlow~\cite{li2024novaflow}, a video-only method that extracts 3D flow from VGM rollouts without constraint-based filtering or refinement.
All methods use the same hardware and perception setup. 
Table~\rev{\ref{tab:success_rekep_novaflow_ours}} reports success counts out of 10 trials per task. The proposed \method{}: 

\textit{i) Achieves substantial improvements over both baselines across all six tasks, improving the overall success rate from 21.7\% (ReKep) and 25.0\% (NovaFlow) to 68.3\%.}
The largest gains appear on tasks requiring precise contact geometry:
Press the Stapler improves by \rev{80 percentage points} over NovaFlow (8/10 vs.\ 0/10) and \rev{60} over ReKep (8/10 vs.\ 2/10), and Place the Block Securely improves by \rev{40 percentage points} over NovaFlow (8/10 vs.\ 4/10) and \rev{70} over ReKep (8/10 vs.\ 1/10).
These improvements stem from our two-stage constraint alignment design: constraint-guided selection filters rollouts with incorrect approach directions or contact locations \emph{before} execution, and constraint-based trajectory optimization refines the retargeted trajectory to satisfy precise spatial requirements, neither of which is available in NovaFlow or ReKep;

\textit{ii) Demonstrates the complementary benefit of combining video proposals with compositional constraints.}
Compared to NovaFlow, which relies solely on VGM rollouts without any constraint filtering, our method adds constraint-guided selection and trajectory optimization, consistently converting physically implausible video plans into executable trajectories.
Compared to ReKep, which plans directly from constraints without video guidance, our method provides VGM-generated rollouts as warm-start initializations for the optimizer, avoiding the local-optima problem that causes ReKep to fail on tasks with complex motion requirements (e.g., Hammer 1/10 $\to$ 4/10, Pour 2/10 $\to$ 7/10).
ReKep's constraint-only optimizer is sensitive to initialization quality, and without a video-guided trajectory prior, it frequently converges to infeasible solutions, particularly on safety-constrained tasks like Place the Block Securely (1/10), where the obstacle creates a non-convex feasible region that a heuristic initial plan cannot navigate.

These results confirm the central thesis of our approach: video proposals provide rich motion priors that address the initialization sensitivity of constraint-only methods, while compositional constraints correct the physical implausibility of video-only pipelines.
Unifying both within a single framework yields consistent improvements across all tasks.

\subsection{Qualitative Results}
\begin{figure}[!t]
  \centering
  \includegraphics[width=0.85\linewidth]{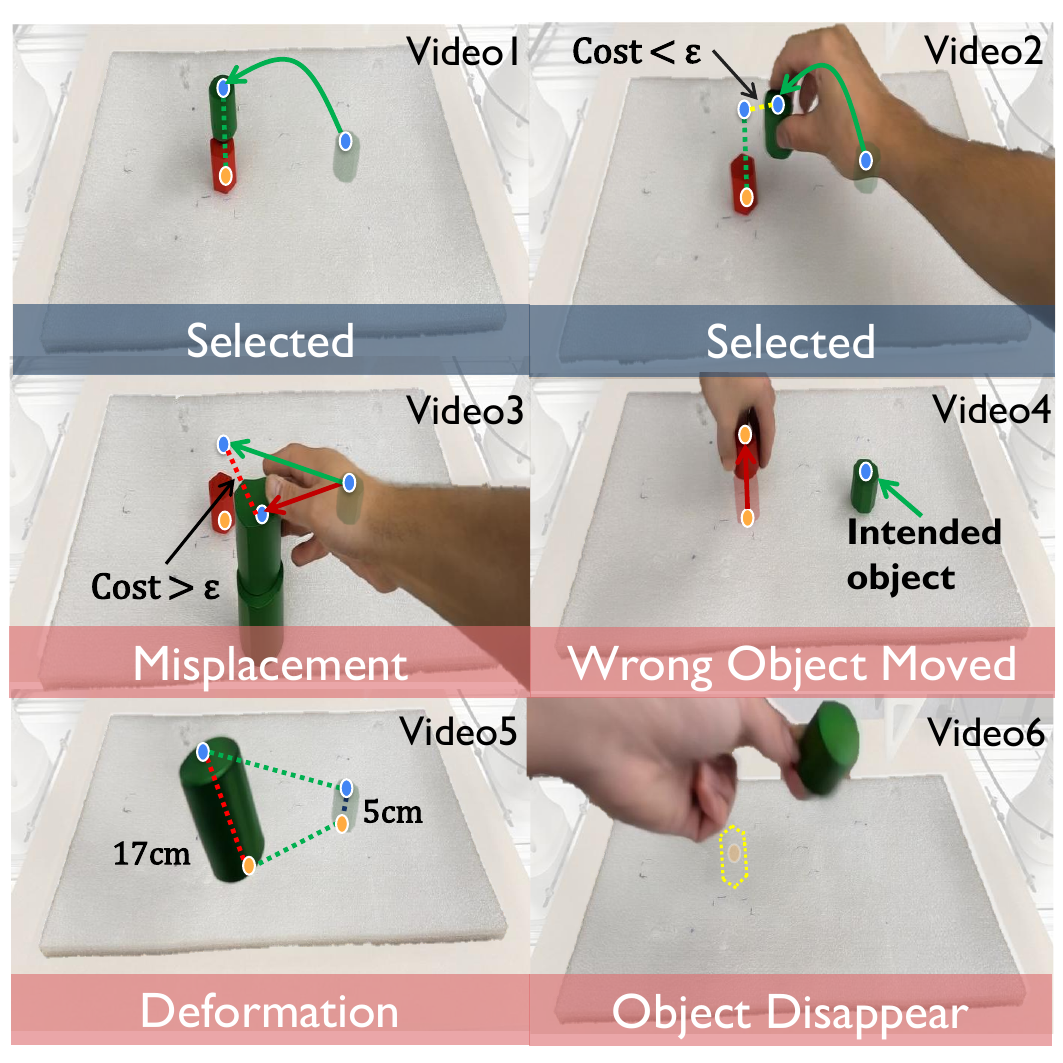}
\caption{\textbf{Constraint-based video selection.} For the \emph{stack} task (place the green block on top of the red block), task constraints filter candidate VGM rollouts by rejecting invalid behaviors. We show representative rejected rollouts and example candidates that \emph{pass all constraints}, \rev{one of which is selected} for downstream retargeting and execution.}
  \label{fig:selection}
  \vspace{-5mm}
\end{figure}

As illustrated in Fig.~\ref{fig:selection}, we visualize the
constraint-based video selection process on the block stacking task. Given six
candidate rollouts generated by the VGM, our constraint checker evaluates each
against the full constraint set and identifies four distinct failure modes:
Video~5 exhibits object deformation where the block shape changes unnaturally
during manipulation, Video~6 shows object disappearance with the red block
vanishing mid-sequence, Video~4 reveals misplacement with the green block
landing at an offset position that violates the spatial alignment constraint,
and Video~3 involves the wrong object being moved as the red block is displaced
instead of remaining static. Only Video~1 and Video~2 satisfy all constraints
with cost below the threshold~$\epsilon$, and the pipeline selects one of them for downstream retargeting and execution. This filtering
step is especially critical for contact-sensitive tasks such as \textit{Press
the Stapler} and \textit{Place the Block Securely}, which show the largest
improvements, underscoring the importance of constraint-guided filtering for contact-sensitive manipulation.

\subsection{Ablation Study}
We conduct \rev{two sets of} ablation studies to isolate the contributions of (i) the underlying video generative model (VGM) used to propose candidate rollouts and (ii) the key components of our pipeline: \emph{video generation} and \emph{constraint-based validation}.
First, we compare three VGMs under the same downstream retargeting and optimization settings, reporting task-level performance in Table~\ref{tab:ablation_vgm}.
Second, we evaluate component-level variants to quantify how constraints and video proposals complement each other: 
(1) \textbf{Constraints-only}, which removes video generation and directly plans using constraints from the initial observation; (2) \textbf{Video-only}, which uses VGM proposals without applying constraint-based filtering; (3) \textbf{+Selection} (Video + Selection), which adds constraint-guided video selection but skips trajectory optimization, executing the selected retargeted trajectory as-is; and (4) \textbf{+Opt} (our full method), which further applies constraint-based trajectory optimization on the selected rollout. Results are summarized in Table~\ref{tab:ablation_components}.
Overall, the VGM comparison highlights the sensitivity to rollout quality, while the component ablations show that constraints are essential for rejecting non-physical or task-inconsistent videos and that constraint-guided selection substantially improves success over using video proposals alone.

\subsection{Failure Mode Analysis}
\begin{figure}[t]
  \centering
  \includegraphics[width=\linewidth]{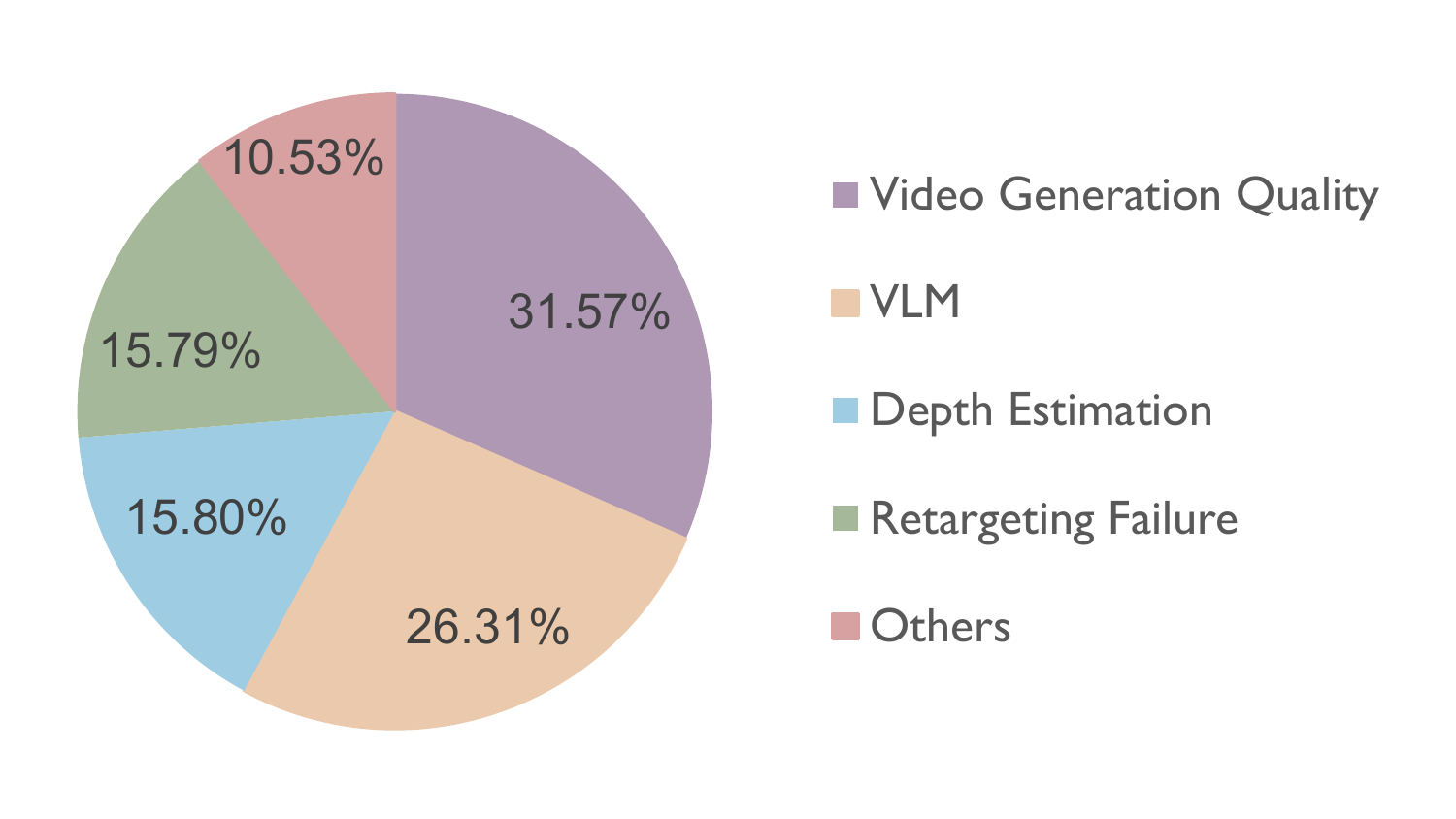}
  \caption{\textbf{Failure mode breakdown.} Distribution of failure causes across all unsuccessful trials.}
  \label{fig:failure}
    \vspace{-2mm}
\end{figure}

We analyze all unsuccessful trials of our full system and categorize them
into five failure modes (Fig.~\ref{fig:failure}).
(1)~\emph{Video generation quality} (31.57\%): the VGM produces rollouts
with subtle physical artifacts (e.g., implausible contact sequences or
inconsistent object dynamics) that pass the constraint filter but lead
to execution failure.
(2)~\emph{VLM keypoint referring} (26.31\%): the VLM keypoint referring failures primarily arise when numbered keypoint labels are spatially proximate or overlapping in the annotated image, causing the VLM to misread or confuse adjacent indices. This is exacerbated in cluttered scenes where multiple objects occupy a small region, leading to incorrect constraint instantiation.
(3)~\emph{Retargeting failure} (15.79\%): errors in keypoint tracking
or rigid-body fitting accumulate during motion retargeting, yielding
infeasible end-effector trajectories that cause missed contacts or
collisions.
(4)~\emph{Depth estimation} (15.80\%): inaccuracies in monocular depth
prediction introduce systematic bias in the 3D keypoint reconstruction,
causing the constraint checker to misjudge spatial relations and the
retargeted trajectory to deviate from the intended motion.
(5)~\emph{Others} (10.53\%): remaining failures include grasp estimation
errors, overly conservative constraint thresholds, and edge-case scene
configurations.

\begin{table}[t]
\centering
\caption{\textbf{Ablation: Effect of the Video Generative Model (VGM).} End-to-end success counts out of 10 trials per task using the same downstream pipeline, varying only the VGM.}
\label{tab:ablation_vgm}
\renewcommand{\arraystretch}{1.15}
\begin{tabularx}{\linewidth}{@{}
    >{\centering\arraybackslash}m{2.2cm}
    >{\centering\arraybackslash}X
    >{\centering\arraybackslash}X
    >{\centering\arraybackslash}X
    >{\centering\arraybackslash}X
    >{\centering\arraybackslash}X
    >{\centering\arraybackslash}X
@{}}
\toprule
\textbf{VGM} &
\textbf{Stack} &
\textbf{Press} &
\textbf{Ham.} &
\textbf{Place} &
\textbf{Open} &
\textbf{Pour} \\
\midrule

\textbf{Wan2.2\cite{wan2025wanopenadvancedlargescale}}     & 5/10 & 7/10 & 2/10 & 6/10 & \textbf{7/10} & 6/10 \\
\textbf{Cosmos2.5\cite{nvidia2026worldsimulationvideofoundation}}  & 4/10 & 6/10 & 3/10 & 6/10 & \textbf{7/10} & 6/10 \\
\textbf{LVP\cite{chen2024lvp}}        & \textbf{7/10} & \textbf{8/10} & \textbf{4/10} & \textbf{8/10} & \textbf{7/10} & \textbf{7/10} \\
\bottomrule
\end{tabularx}
\end{table}

\begin{table}[t]
\centering
\caption{\textbf{Ablation: Video vs. Constraints.} Success counts out of 10 trials per task.}
\label{tab:ablation_components}
\renewcommand{\arraystretch}{1.05}
\setlength{\tabcolsep}{3.5pt}
\begin{tabularx}{\linewidth}{@{}
    >{\centering\arraybackslash}m{2.1cm}
    >{\centering\arraybackslash}X
    >{\centering\arraybackslash}X
    >{\centering\arraybackslash}X
    >{\centering\arraybackslash}X
    >{\centering\arraybackslash}X
    >{\centering\arraybackslash}X
    >{\centering\arraybackslash}m{1.0cm}
@{}}
\toprule
\textbf{Variant} &
\textbf{Stack} &
\textbf{Press} &
\textbf{Ham.} &
\textbf{Place} &
\textbf{Open} &
\textbf{Pour} &
\textbf{Avg.} \\
\midrule
\textbf{Constr.-only} & 2/10 & 3/10 & 2/10 & 2/10 & 5/10 & 3/10 & 28.3\% \\
\textbf{Video-only}   & 3/10 & 1/10 & 0/10 & 3/10 & 4/10 & 3/10 & 23.3\% \\
\textbf{+ Selection}  & 5/10 & 6/10 & 2/10 & 6/10 & 5/10 & 5/10 & 48.3\% \\
\textbf{+ Opt} &  \textbf{7/10} & \textbf{8/10} & \textbf{4/10} & \textbf{8/10} & \textbf{7/10} & \textbf{7/10} & \textbf{68.3\%} \\
\bottomrule
\end{tabularx}

\vspace{2pt}
{\footnotesize \textbf{Constr.-only}: no VGM proposals; plans directly from constraints.
\textbf{Video-only}: no constraint filtering or trajectory refinement. All video-based variants use LVP~\cite{chen2024lvp} as the default VGM.}
  \vspace{-5mm}
\end{table}

\section{CONCLUSIONS}

We present \method{}, a framework that aligns video generative models with compositional physical constraints for zero-shot robotic manipulation.
By leveraging the complementarity between VGMs (rich motion priors) and VLMs (structured constraint reasoning), \method{} applies VLM-derived constraints at two stages, rollout selection and trajectory optimization, to address physical hallucinations and retargeting errors without modifying any pretrained model weights.
Experiments on six real-robot tasks demonstrate a 68.3\% average success rate, with substantial gains over both video-only and constraint-only baselines.





\section*{ACKNOWLEDGMENT}

We graciously acknowledge the support from National Science Foundation grants 2324936 and 2328973.


\bibliographystyle{ieeetr}
\bibliography{references}

\end{document}